\documentclass[acmtog]{acmart}

\renewcommand\footnotetextcopyrightpermission[1]{} 

\setcopyright{rightsretained}
\acmJournal{TOG}
\acmYear{2020}
\acmVolume{39}
\acmNumber{6}
\acmArticle{257}
\acmMonth{12} 
\acmDOI{10.1145/3414685.3417827}

\setcitestyle{square}

\usepackage{soul}
\usepackage{microtype}
\usepackage{amsmath}
\usepackage{amsfonts}
\usepackage{booktabs}
\usepackage{hyperref}
\usepackage{multirow}
\usepackage{graphicx}
\usepackage{dsfont}
\usepackage{pifont}
\usepackage{xcolor}
\usepackage{tikz}
\usepackage{wrapfig}
\usepackage{nicefrac}
\usepackage{bbm}
\usepackage{flushend}
\usepackage{ctable}
\usepackage{natbib}
\usepackage{xspace}

\usepackage{soul}
\usepackage{amsmath}
\usepackage{microtype}
\usepackage{wrapfig}

\def\figurePath{images/}
\def\myfigure#1#2{\begin{figure}[htb]\centering\includegraphics*[width = \linewidth]{\figurePath#1}\vspace{-0.1cm}\caption{#2}\vspace{-0.25cm}\label{fig:#1}\end{figure}}
\def\mycfigure#1#2{\begin{figure*}[t]\centering\includegraphics*[clip, width = \linewidth]{\figurePath#1}\vspace{-0.1cm}\caption{#2}\vspace{-0.25cm}\label{fig:#1}\end{figure*}}
\newcommand{\mywrapfigure}[3]{%
\begin{wrapfigure}{r}{#2\columnwidth}%
  \begin{center}%
    \includegraphics[width=#2\columnwidth]{\figurePath#1}%
    \vspace{-0.3cm}%
    \caption{#3}%
    \label{fig:#1}%
  \end{center}%
\end{wrapfigure}%
\leavevmode%
}

\newcommand{\eg}{e.g., }
\newcommand{\ie}{i.e., }
\newcommand{\etal}{et~al.\ }

\newcommand{\citeetal}[1]{et~al.~\shortcite{#1}}
\newcommand{\namecite}[1]{\citeauthor{#1}~\shortcite{#1}}

\newcommand{\argmin}[1]{\underset{#1}{\operatorname{arg\,min}\ }}

\newcommand{\refSec}[1]{Sec.~\ref{sec:#1}}
\newcommand{\refFig}[1]{Fig.~\ref{fig:#1}}
\newcommand{\refEq}[1]{Eq.~\ref{eq:#1}}
\newcommand{\refTbl}[1]{Tbl.~\ref{tbl:#1}}
\newcommand{\refFigStart}[1]{Figure~\ref{fig:#1}}

\soulregister\ref7
\soulregister\cite7
\soulregister\refFig7
\soulregister\refFigStart7
\soulregister\cite7
\soulregister\ref7
\soulregister\pageref7
\soulregister\shortcite7
\soulregister\namecite7
\soulregister\eg7
\soulregister\ie7
\soulregister\etal7
\soulregister\reftbl7
\soulregister\name7
\soulregister\unsure7
\soulregister\url7

\DeclareGraphicsExtensions{.png,.jpg,.pdf,.ai,.psd}
\newcommand{\mysection}[2]{\section{#1}\label{sec:#2}}
\newcommand{\mysubsection}[2]{\subsection{#1}\label{sec:#2}}
\newcommand{\mysubsubsection}[2]{\subsubsection{#1}\label{sec:#2}}

\def\figurePath{images/}

\newcommand{\name}{X-Field\xspace}
\newcommand{\names}{X-Fields\xspace}

\newcommand{\xFieldOut}{L_\mathrm{out}}
\newcommand{\xFieldIn}{L_\mathrm{in}}
\newcommand{\slice}{I}
\newcommand{\slices}{\mathcal I}
\newcommand{\xCoord}{\mathbf x}
\newcommand{\xCoords}{\mathcal X}
\newcommand{\dimension}{{n_\mathrm d}}
\newcommand{\numberOfPixels}{{n_\mathrm p}}
\newcommand{\parameters}{\theta}
\newcommand{\expected}{\mathbb E}
\newcommand{\sparseXCoords}{\mathcal Y}
\newcommand{\sparseXCoord}{\mathbf y}
\newcommand{\shading}{E}
\newcommand{\albedo}{A}
\newcommand{\interpolate}{\mathtt{int}}
\newcommand{\pixelPosition}{\mathbf{p}}
\newcommand{\warpedPixelPosition}{\mathbf{q}}
\newcommand{\flow}{\mathtt{flow}_\partial}
\newcommand{\flowPix}{\mathtt{flow}_\Delta}
\newcommand{\warp}{\mathtt{warp}}
\newcommand{\consistency}{\mathtt{cons}}

\newcommand{\consistencyWeight}{w}
\newcommand{\bandwidth}{\sigma}
\newcommand{\heldOutXCoords}{\mathcal H}
\newcommand{\heldOutXCoord}{\mathbf h}

\def\myfigure#1#2{\begin{figure}[ht]\centering\includegraphics*[width = \linewidth]{\figurePath#1}\vspace{-.2cm}\caption{#2}\label{fig:#1}\end{figure}}

\def\mycfigure#1#2{\begin{figure*}[t]\centering\includegraphics*[clip, width = \linewidth]{\figurePath#1}\vspace{-.2cm}\caption{#2}\label{fig:#1}\end{figure*}}

\def\mysection#1#2{\section{#1}\label{sec:#2}}
\def\mysubsection#1#2{\subsection{#1}\label{sec:#2}}
\def\mysubsubsection#1#2{\subsubsection{#1}\label{sec:#2}}

\definecolor{fixedcolor}{rgb}{.8,1,.7}
\definecolor{fixedncolor}{rgb}{.8,.2,.1}
\definecolor{revisioncolor}{rgb}{1,.95,.4}
\definecolor{revisionfinalcolor}{rgb}{0.3,1,0.76}

\DeclareRobustCommand{\change}[1]{#1}

\DeclareRobustCommand{\revision}[1]{#1}

\soulregister\ref7
\soulregister\cite7
\soulregister\citeetal7
\soulregister\refSec7
\soulregister\refFig7
\soulregister\refTbl7
\soulregister\refEq7
\soulregister\cite7
\soulregister\ref7
\soulregister\pageref7
\soulregister\shortcite7
\soulregister\eg7
\soulregister\ie7
\soulregister\etal7
\soulregister\unsure7

\DeclareGraphicsExtensions{.png,.jpg,.pdf,.ai,.psd}
\usepackage{pifont}
\newcommand{\cmark}{\checkmark}%
\newcommand{\xmark}{\scalebox{0.85}{\ding{53}}}%
\usepackage{adjustbox}
\newcolumntype{R}{%
    >{\adjustbox{angle=90}\bgroup}%
    l%
    <{\egroup}%
}
\newcommand*\rot{\multicolumn{1}{R}}%

\newcommand{\resultMethod}[2]{
    \multicolumn{3}{c}{
        {\color[HTML]{#2}\ding{108}}
        \textsc{#1}
    }
}

\newcommand{\ndcell}{\multicolumn{1}{c}{---}}
\usepackage{stringstrings}
\usepackage{forloop}

\newcommand{\methodFootnote}[1] {%
    \stepcounter{methodFootNoteCounter}%
    #1%
    $^{\themethodFootNoteCounter}$%
}

\newcommand{\method}[4]{
    #2\hphantom{A}&%
    \def\temp{#3}\ifx\temp\empty
    \else
        \citeauthor{#3} \shortcite{#3}\hphantom{A}%
    \fi
    \setcounter{foo}{1}
    \forloop{foo}{1}{\value{foo} < 12}{
        &%
        \StrChar{#1}{\thefoo}[\pfft]%
        \IfStrEq
            {\pfft}
            {c}
            {\cmark\hphantom{1}}{}%
        \IfStrEq
            {\pfft}
            {x}
            {\xmark\hphantom{1}}{}%
        \IfStrEq
            {\pfft}
            {X}
            {\methodFootnote{\xmark}}{}%
        \IfStrEq
            {\pfft}
            {C}
            {\methodFootnote{\cmark}}{}%
    }%
    \hphantom{A}&#4\\
}

\begin{document}

\newcounter{foo}
\newcounter{methodFootNoteCounter}

\title{X-Fields: Implicit Neural View-, Light- and Time-Image Interpolation}

\author{Mojtaba Bemana}
\affiliation{%
	\institution{MPI Informatik, Saarland Informatics Campus}
}
\email{mbemana@mpi-inf.mpg.de}

\author{Karol Myszkowski}
\affiliation{%
	\institution{MPI Informatik, Saarland Informatics Campus}
}

\author{Hans-Peter Seidel}
\affiliation{%
	\institution{MPI Informatik, Saarland Informatics Campus}
}

\author{Tobias Ritschel}
\affiliation{%
	\institution{University College London}
}

\begin{teaserfigure}
   \includegraphics[width=\textwidth]{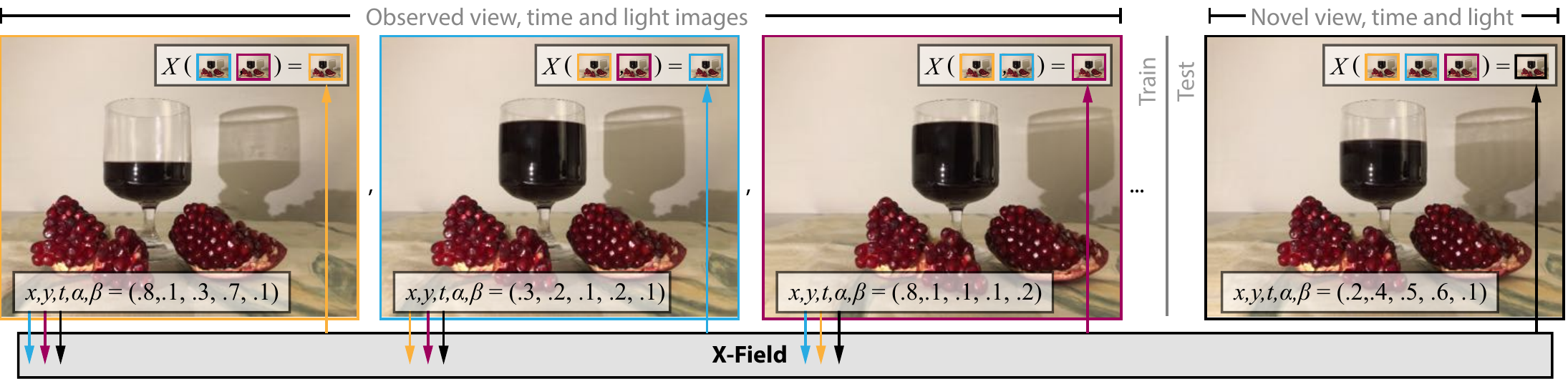}
   \vspace{-.4cm}
   \caption{
   To interpolate view, light and time in a set of 2D images labeled with coordinates \change{(\name)}, we train a \change{neural network (NN)} to regress each image from all others.
   The first (yellow) image is the NN output (yellow up arrow) when the blue and purple observed images and their coordinates $x,y,t,\alpha,\beta$ are input (yellow down arrows).
   The blue and purple observations form additional constraints, visualized as colored boxes.
   Provided with an unobserved coordinate (black up arrow) the NN produces, from the observed images and coordinates (black down arrow), a novel high-quality 2D image in real time.
   }
   \label{fig:Teaser}
\end{teaserfigure}

\begin{abstract}
We suggest to represent an \name \textemdash a set of 2D images taken across different view, time or illumination conditions, \ie video, lightfield, reflectance fields or combinations thereof\textemdash by learning a neural network (NN) to map their view, time or light coordinates to 2D images.
Executing this NN at new coordinates results in joint view, time or light interpolation.
The key idea to make this workable is a NN that  already knows the ``basic tricks'' of graphics (lighting, 3D projection, occlusion) in a hard-coded and differentiable form.
The NN represents the input to that rendering as an implicit map, that for any view, time, or light coordinate and for any pixel can quantify how it will move if view, time or light coordinates change (Jacobian of pixel position with respect to view, time, illumination, etc.).
Our \revision{\name\ representation} is trained for one scene within minutes, leading to a compact set of trainable parameters and hence real-time navigation in view, time and illumination.
\end{abstract}

\maketitle

\thispagestyle{empty}
\pagestyle{plain}

\mysection{Introduction}{Introduction}
Current and future sensors capture images of one scene from different points (video), from different angles (light fields), under varying illumination (reflectance fields) or subject to many other possible changes.
In theory, this information will  allow exploring time, view or light changes in Virtual Reality (VR).
Regrettably, in practice, sampling this data densely leads to excessive storage, capture and processing requirements.
In higher dimensions\textemdash here we demonstrate 5D\textemdash the demands of dense regular sampling (cubature) increase exponentially.
Alternatively, sparse and irregular sampling overcomes these limitations, but requires faithful interpolation across time, view and light.
We suggest taking an abstract view on all those dimensions and simply denote any set of images conditioned on parameters as an ``\name'', where X could stand for any combination of time, view, light or other dimensions like the color spectrum.
We will demonstrate how the right neural network (NN) becomes a universal, compact and interpolatable \name representation.

While NNs have been suggested to estimate depth or correspondence across space, time or light, we here, for the first time, suggest representing the complete \name implicitly \cite{niemeyer2019occupancy,chen2019implicit,oechsle2019texturefields,sitzmann2019srns}, \ie as a trainable architecture that implements a high-dimensional \texttt{getPixel}.
The main idea is shown in \refFig{Teaser}: from sparse image observations with varying conditions and coordinates, we train a mapping that, when provided the space, time or light coordinate as an input, generates the observed sample image as an output.
Importantly, when given a non-observed coordinate, the output is a faithfully interpolated image. 
Key to making this work is the right training and a suitable network structure, involving a very primitive (but differentiable) rendering (projection and lighting) step.

Our architecture is trained for one specific \name to generalize across its parameters, but not across scenes.
However, per-scene training is fast (minutes), and decoding occurs at high frame rates (ca.\ 20~Hz) and high resolution (1024$\times$1024).
In a typical use case of the VR exploration of an \name, the architecture parameters only require a few additional kilobytes on top of the image samples.
We compare the resulting quality to several other state-of-the-art interpolation baselines (NN and classic, specific to certain domains and general) as well as to ablations of our approach.

\revision{
Our neural network implementation and training data is publicly available at 
\url{http://xfields.mpi-inf.mpg.de/}.
}

\mysection{Previous Work}{PreviousWork}
Here we review previous techniques to interpolate across discrete sampled observations in view (light fields), time (video) and illumination (reflectance fields).
\refTbl{PreviousWork} summarizes this body of work along \revision{multiple axes}.
\revision{\namecite{tewari2020star} provide further report of the state-of-the-art in Neural Rendering.}

\begin{table*}[h]
    \setlength{\tabcolsep}{0pt}
    \centering
    \caption{
    \change{
    Comparison of space, time, and illumination interpolation methods \emph{(rows)} in respect to  capabilities \emph{(columns)}, with an emphasis on deep methods.  \\
    ($^{1-3,5}$Similar-class scenes demonstrated, \eg cars, chairs, urban city views;
    $^4$LF sparse in time;
    $^6$Clothed humans demonstrated;
    $^7$Human faces shown;
    $^8$\revision{Only structured grids shown. Should support unstructured, as long as the transformation between views are known.})
    }
    }
    \begin{tabular}{llcccccccccccl}
        &
        &
        \rot{Scene}&
        \rot{View}&
        \rot{Time}&
        \rot{Light}&
        \rot{Sparse}&
        \rot{Unstructured}&
        \rot{Real-time}&
        \rot{Easy capture}& 
        \rot{Learned}&
        \rot{Implicit}&
        \rot{Diff.\ render.}
        \\
        \cmidrule(r{.5em}){3-6}  
        \cmidrule(l{1.35em}r{.5em}){6-10}  
        \cmidrule(l{1.35em}r{1.15em}){10-13}
        \multicolumn{1}{c}{Method}&
        \multicolumn{1}{c}{Citation}&
        \multicolumn{4}{l}{Generalize}&
        \multicolumn{4}{c}{Interface}&
        \multicolumn{3}{l}{Implem.}&
        \multicolumn{1}{c}{Remarks}
        \\
        \toprule        
\method{ccxxccxcxxx}
{Unstructured\ Lumigraph}{buehler2001unstructured}{IBR}
\method{ccxxccxcxxx}
{Soft 3D}{penner2017soft}{MPI}
\method{ccxxccccxxx}
{Inside Out}{hedman2016insideout}{IBR; SfM; Per-view geometry}
\method{ccxxcccccxx}
{Deep Blending}{hedman2018blending}{IBR; SfM; Learned fusion}
\method{ccxxxxxccxx}
{Learning-based View Interp.}{Kalantari2016}{Lytro; Learned disparity and fusion}
\method{ccxxcccccxx}
{Local LF Fusion}{Mildenhall2019}{MPI}
\method{xcxxccxxcxx}
{DeepView}{flynn2019deepview}{MPI}
\method{xcxxxxcxcxx}
{Deep Surface LFs}{chen2018deep}{Texture; Lumitexel; MLPs}
\method{xcxxxcxxccc}{\change{NeRF}}{mildenhall2020nerf}{MLPs, ray-marching}
\method{xcxxxxcxccc}
{Neural Textures}{thies2019neural}{Texture; Lumitexel; CNNs}
\method{xcxxxcxxccc}
{DeepVoxels}{sitzmann2019deepvoxels}{3D CNN}
\method{ccxxxxxccxc}
{HoloGAN}{nguyen2019hologan}{Adversarial; 3D representation}
\method{Ccxxxxxxccc}
{Appearance Flow}{zhou2016appearance}{App.\ Flow; Fixed views}
\method{Ccxxxxxxccc}
{Multi-view App. Flow}{sun2018multi}{App.\ Flow; Learned fusion; Fixed views}
\method{Ccxxxccxccc}
{Spatial Trans. Net IBR}{chen2019monocular}{App.\ Flow; Per-view geometry; Free views} 
\midrule
\method{cxcxxxxcxxx}
{Moving Gradients}{mahajan2009moving}{Gradient domain}
\method{cxcxcxxccxx}
{Super SlowMo}{jiang2018super}{Occlusions: Learns visibility maps}
\method{cxcxcxxccxx}
{MEMC-Net}{bao2019memc}{Occlusions: Learns visibility maps}
\method{cxcxcxxccxx}
{Depth-aware Frame Int.}{bao2019depth}{Occlusions: Learns depth maps}
\method{cxcxxxxccxx}
{Video-to-video}{wang2018vid2vid}{Adversarial; Segmented content editing}
\method{xxcxxxccccxx}
{Puppet Dubbing}{ohad2019dubbing}{Visual and sound sync.}
\midrule
\method{cccxcccxxxx}
{Layered Representation}{zitnick2004high}{MVS reconst.; Layered Depth Images  }
\method{cccxxxxxxxx}{\revision{Video Array}}{wilburn2005high}{\revision{Optical flow}}
\method{cccxccccxxx}
{Virtual Video}{lipski10Virtualvideo}{Structure from Motion (SfM) }
\method{cccxCxcccxx}
{Hybrid Imaging}{wang2017light}{Lytro+DSLR camera system }
\method{xccxxxcxccc}{Neural Volumes}{lombardi2019volumes}{3D CNN; Lightstage; Fixed time (video)}
\method{Cccxxxxxccc}{Scene Represent. Net}{sitzmann2019srns}{3D MLP}
\method{Cccxccxcccc}
{Pixel-aligned Implicit Funct.}{saito2019pifu}{3D MLP}
\midrule
\method{xxccxxcxxcx}
{Polynomial Textures}{malzbender2001polynomial}{Lightstage}
\method{xxxcxcxxccx}
{Neural Relighting}{ren2015image}{MLP; Lightstage and hand-held lighting}
\method{cxxccccxcxx}
{Sparse Sample Relighting}{xu2018deeprelighting}{Optimized light positions}
\midrule
\method{ccxccccxcxx}
{Sparse Sample View Synth.}{xu2019deepviewsynth}{Optimized lights as in \cite{xu2018deeprelighting}}
\method{ccxccccccxx}
{Multi-view Relighting}{philip2019}{Geometry proxy; Auxiliary 2D buffers}
\midrule
\method{Ccccxxxxccx}{Deep Reflectance Fields}{meka2019deepreflectance}{Lightstage}
\method{ccccxxcxcxx}{The Relightables}{guo2019relighttables}{Lightstage}
\midrule
\method{xccccXcccccc}
{Ours}{}{}
    \bottomrule
    \end{tabular}
    \label{tbl:PreviousWork}
\end{table*}

\mysubsection{View Interpolation (Light Fields)}{PWView}
\namecite{levoy1996light} as well as  \namecite{gortler1996lumigraph} were first to formalize the concept of a light field (LF)\textemdash the set of all images of a scene for all views\textemdash and to devise hardware to capture it.
LF methods come first in \refTbl{PreviousWork}, where they are  checked ``view'' as they generalize across an observer's position and orientation.
A simple solution for all interpolation, including view, is linear blending, but this leads to ghosting.

An important distinction is that a capture can be \emph{dense} or \emph{sparse}, denoted as ``Sparse'' in \refTbl{PreviousWork}.
Sparsity depends less on the number of images, but more on the difference between captured images. 
Very similar view positions \cite{Kalantari2016} as for a Lytro camera can be considered dense, while 34 views on a sphere \cite{lombardi2019volumes} or 40 lights on a hemisphere \cite{malzbender2001polynomial} is sparse.
In this paper we focus on wider baselines, with typically $M \times N$ cameras spaced by 5--10\,cm \cite{flynn2019deepview}, and respectively a large disparity ranging up to 250~pixels \cite{dabala2016lightfield,Mildenhall2019},
where $M$ and $N$ are single-digit numbers, \eg 3$\times$3, 5$\times$5 or even 2$\times$1.
\change{Depending on resources, a capture setup can be considered simple (cell phone, as we use) or more involved (light stage) as denoted in the ``easy capture'' column in \refTbl{PreviousWork}.}

Early view interpolation solutions, such as Unstructured Lumigraph Rendering (ULR) \cite{buehler2001unstructured,Chaurasia2013}, typically create proxy geometry to warp \cite{mark1997post} multiple observations into a novel view and blend them with specific weights.
More recent work has used per-view geometry \cite{hedman2016insideout} and learned ULR blending weights \cite{hedman2018blending}, to allow sparse input and view-dependent shading.
Avoiding the difficulty of reconstructing geometry or 3D volumes has been addressed for LFs in \cite{DuDDHM2013,Kellnhofer2017}.

An attractive recent idea is to learn synthesizing novel views for LF data. 
\namecite{Kalantari2016} indirectly learn depth maps without depth supervision to interpolate between views in a Lytro camera. 
Another option is to decompose LFs into multiple depth planes of the output view and construct a view-dependent plane sweep volume (PSV) \cite{flynn2016deepstereo}.
By learning how neighboring input views contribute to the output view, the multi-plane image (MPI) representation \cite{Zhou2018StereoMagnification} can be built, which enables high-quality local LF fusion \cite{Mildenhall2019}.

Instead of using proxy geometry, \namecite{penner2017soft} have suggested using a volumetric occupancy representation.
Inferring a good volumetric / MPI representation can be facilitated with learned gradient descent \cite{flynn2019deepview}, where the gradient components directly encode visibility and effectively inform the NN on the occlusion relations in the scene.
MPI techniques avoid the problem of explicit depth reconstruction and allow for softer, more pleasant results.
A drawback in deployment is the massive volumetric data, the difficulty of distributing occupancy therein, and finally bandwidth requirements of volume rendering itself.
Our approach involves a learning route as well, but explaining the entire \name and using a NN to represent the scene implicitly.
Deployment only requires a few additional kilobytes of NN parameters on top of the input image data, and rendering is real-time.

From yet another angle, ULR-inspired IBR creates a LF (view-dependent appearance) on the surface of a proxy geometry, \ie a \emph{surface light field}.
\namecite{chen2018deep}, using an MLP, as well as \namecite{thies2019neural}, using a CNN, have proposed to represent this information using a NN defined in texture space of a proxy object.
While inspired by the mechanics of sparse IBR, results are typically demonstrated for rather dense observations.
Our approach does not assume any proxy to be given, but jointly represents the appearance and the geometry used to warp over many dimensions, in a single NN, trained from sparse sets of images.

Another step of abstraction is Deep Voxels \cite{sitzmann2019deepvoxels}.
Instead of storing opacity and appearance in a volume, abstract ``persistent'' features are derived, which are projected using learned ray-marching.
The volume was learned per-scene.
\change{In this work we also use implicit functions instead of voxels, as did \namecite{sitzmann2019srns}, \namecite{saito2019pifu} and \namecite{mildenhall2020nerf}.} 
We call such approaches ``implicit'' in \refTbl{PreviousWork} when the NN replaces the pixel basis, \ie the network provides a high-dimensional \texttt{getPixel(x)}.
These approaches use an MLP that can be queried for occupancy \cite{chen2019implicit,sitzmann2019srns,saito2019pifu}, color \cite{oechsle2019texturefields,sitzmann2019srns,mildenhall2020nerf}, flow \cite{niemeyer2019occupancy} etc. at different 3D positions along a ray for one pixel.
\change{We make two changes to this design.
First, we predict texture coordinates, rather than appearance.
These drive a spatial transformer \cite{jaderberg2015spatial} that can copy details from the input images without representing them and do so at high speed (20~fps).
Second, we train a 2D CNN instead of a 3D MLP that, for a given \name{} coordinate, will directly return a complete 2D per-pixel depth and correspondence map.
For an \name{} problem, this is more efficient than ray-marching and evaluating a complex MLP at every step \cite{oechsle2019texturefields,niemeyer2019occupancy,sitzmann2019srns,mildenhall2020nerf}.
}
While implicit representations have so far been demonstrated to provide a certain level of fidelity when generalized across a class of simpler shapes (cars, chairs, etc.), we here make the task simpler and generalize less, but produce quality to compete with state-of-the-art view, time and light interpolation methods from computer graphics.

Inspired by \namecite{nguyen2018rendernet}, some work \cite{sitzmann2019deepvoxels,nguyen2019hologan,sitzmann2019srns} learns the differentiable tomographic rendering step, while other work has shown how it can be differentiated directly \cite{henzler2019platonicgan,lombardi2019volumes,mildenhall2020nerf}.
Our approach avoids tomography and works with differentiable warping
\cite{jaderberg2015spatial} with consistency handling inspired by unsupervised depth reconstruction \cite{godard2017monodepth,zhou2017unsupervised}.
Avoiding volumetric representations allows for real-time playback, while at the same generalizing from view to other dimensions such as time and light.

The Appearance Flow work of \namecite{zhou2016appearance} suggests to combine the idea of warping pixels with learning how to warp.
While \namecite{zhou2016appearance} typically consider a single input view, \namecite{sun2018multi} employ multiple views to improve warped view quality.
Both works use an implicit representation of the warp field, \ie a NN that for every pixel in one view predicts from where to copy its value in the new view.
While those techniques worked best for fixed camera positions that are used in training, \namecite{chen2019monocular} introduces an implicit NN of per-pixel depth that enables an arbitrary view interpolation.
All these methods require an extensive training for specific classes of scenes such as cars, chairs, or urban city views.
We take this line of work further by constructing an implicit NN representation that generalizes jointly over complete geometry, motion, and illumination changes.
Our task on the one hand is simpler, as we do not generalize across different scenes, yet on the other hand it is also harder, as we generalize across many more dimensions and provide state-of-the-art visual quality.

\mysubsection{Time (Video)}{PWTime}
Videos comprise discrete observations, and hence are also a sparse capture of the visual world.
To get smooth interpolation, \eg for slow-motion (individual frames), motion blur (averaging multiple frames) images need to be interpolated \cite{mahajan2009moving}, potentially using NNs \cite{sun2018pwc,jiang2018super,bao2019memc,bao2019depth,wang2018vid2vid}.
More exotic domains of video re-timing, which involve annotation of a fraction of frames and one-off NN training, include the visual aspect in sync with spoken language \cite{ohad2019dubbing}.



\mysubsection{Space-Time}{PWMultiple}
Warping can be applied to space or time, as well as to both jointly \cite{manning1999interpolating}, resulting in LF video \cite{wang2005towards,lipski10Virtualvideo,wang2017light,zitnick2004high}.

Recent work has extended deep novel-view methods into the time domain \cite{lombardi2019volumes}, and is closest to our approach.
They also use warping, but for a very different purpose: deforming a pixel-basis 3D representation over time in order to avoid storing individual frames (motion compensation).
Both methods \namecite{sitzmann2019deepvoxels} and \namecite{lombardi2019volumes} are limited by the spatial 3D resolution of volume texture and the need to process it, while we work in 2D depth and color maps only.
As they learn the tomographic operator, this limit in resolution is not a classic Nyquist limit, \eg sharp edges can be handled, but results typically are on isolated, dominantly convex objects, while we target entire scenes.
Ultimately, we do not claim depth maps to be superior to volumes per se.
Instead, we suggest that 3D volumes have their strength for seeing objects from all views (at the expense of resolution), whereas our work, using images, is more for observing scenes from a ``funnel'' of views, but at high 2D resolution.
No work yet is able to combine high resolution and arbitrary views, not to mention time.

\mysubsection{Light Interpolation (Reflectance Fields)}{PWLight}
While a LF is specific to one illumination, a \emph{reflectance field} (RF) \cite{debevec2000acquiring} is a generalization additionally allowing for relighting, often just for a fixed view.
Dense sampling for individually controlled directional lights can be performed using Lightstage \cite{debevec2000acquiring}, which leads to hundreds of captured images.
The number of images can be reduced by employing specially designed illumination patterns \cite{fuchs2007,peers2009,reddy2012} to exploit various forms of coherence in the light transport function.
Our capture is from uncalibrated sets of flash images of mobile phones.
For interpolation, the signal is frequently separated, such as into highlights, reflectance or shadows \cite{chen2005light}.
We also found such a separation to help.
Angular coherence in incoming lighting leads to an efficient reflectance field representation as polynomial texture maps  \cite{malzbender2001polynomial}, which can be further improved by neural networks whose expressive power enables one to capture non-linear spatial coherence \cite{ren2015image}, or generalize across views \cite{maximov2019dam}.
\namecite{xu2018deeprelighting} directly regresses images of illumination from an arbitrary light direction when given five images from specific other light directions.
The innovation is in optimizing what should be input at test time, but the setup requires custom capture dome equipment, as well as input images taken from those five, very specific, directions.
For scenes captured under controlled illumination for multiple sparse views, generalization across views can be achieved by concatenation with a view synthesis method \cite{xu2019deepviewsynth}.
While the results are compelling on synthetic scenes, the method exhibits difficulties in handling complex or non-convex geometry, as well as high frequency details such as specularities and shadows \cite{meka2019deepreflectance}. 
An approximate geometry proxy and extensive training over rendered scenes might compensate for inaccuracies in derived shadows and overall relighting quality \cite{philip2019}.

Specialized systems for relighting human faces and characters remove many such limitations, including fixed view and static scene assumptions, using advanced Lightstage hardware that enables capturing massive data for CNN training \cite{meka2019deepreflectance} and complex optimizations that are additionally fed with multiple depth sensors' data \cite{guo2019relighttables}.
As only two images for an arbitrary face or character under spherical color gradient lighting are required at the test time, real-time dynamic performance capturing is possible.
CNN-based, LF-style view interpolation is performed in  \namecite{meka2019deepreflectance}, whereas \namecite{guo2019relighttables} capture complete 3D models with textures and can easily change viewpoint as well.

 \namecite{meka2019deepreflectance} and \namecite{guo2019relighttables} generalize over similar scenes (faces) while our approach is fixed to one specific scene. 
On the other hand, we remove the requirements for massive training data and costly capturing hardware, while our lightweight network enables real-time rendering of animated scenes under interpolated dynamic lighting and view position.

\mysection{Background}{Background}

Two main observations motivate our approach: First, representing information using NNs leads to interpolation.
Second, this property is retained, if the network contains more useful layers, such as a differentiable rendering step.
Both will be discussed next:

\myfigure{NNInterpolation}{NN and pixel interpolation:
\textbf{a)} Flatland interpolation in the pixel \textbf{(lines)} and the NN representations \textbf{(dotted lines)} compared to a reference \textbf{(solid)} for a 1D field (vertical axis angle; horizontal axis space).
The top and bottom are observed and the middle is unobserved, \ie interpolated.
\textbf{b,c)} Comparing the continuous interpolation in the pixel and the NN representation visualized as a (generalized) epi-polar image.
Note that the NN leads to smooth interpolation, while the pixel representation causes undesired fade-in/fade-out transitions.
}
\paragraph{Deep representations help interpolation.}
It is well-known that deep representations suit interpolation of 2D images \cite{radford2015unsupervised,reed2015deep,white2016sampling}, audio \cite{engel2017neural} or 3D shape \cite{dosovitskiy2015learning} much better than the pixel basis.

Consider the blue and orange bumps in \refFig{NNInterpolation}, a; these are observed.
They represent flat-land functions of appearance (vertical axis), depending on some abstract domain (horizontal axis), that later will become space, time, reflectance etc. in an \name.
We wish to interpolate something similar to the unobserved violet bump in the middle.
Linear interpolation in the pixel basis (solid lines) will fade both in, resulting in two flat copies.
Visually this would be unappealing and distracting ghosting.
This difference is also seen in the continuous setting of \refFig{NNInterpolation}, b that can be compared to the reference in \refFig{NNInterpolation}, c.
When representing the bumps as NNs to map coordinates to color (dotted lines), we note that they are slightly worse than the pixel basis and might not match the NNs.
However, the interpolated, unobserved result is much closer to the reference, and this is what matters in \name interpolation.

To benefit from interpolation, typically, substantial effort is made to construct latent codes from images, such as auto-encoders \cite{hinton2006reducing}, variational auto-encoders  \cite{kingma2013auto} or adversarial networks \cite{goodfellow2014generative}.
We make the simple observation that this step is not required in the common graphics task of image (generalized) interpolation.
In our problem we already have the latent space given as beautifully laid-out space-time \name coordinates and only need to learn to decode these into images.

\myfigure{Epipolar}{\revision{Validation experiment:
Different interpolation \textbf{(rows)}, for two variants \textbf{(columns)} of a right-moving SIGGRAPH Asia 2020 logo (a 1D \name).
For each method we show the same epipolar slice (\ie space on the horizontal axis; time on the vertical axis) marked in the input image.
Nearest and linear sampling show either blur or step artifacts.
A NN to interpolate solid color depending on time succeeds, but lacks capacity to reproduce textured details, where the fine diagonal stripes are missing.
A NN to interpolate flow instead, also captures the textured stripes.}}

\paragraph{(Differentiable) rendering is just another non-linearity.}
The second key insight is that the above property holds for any architecture as long as all units are differentiable.
In particular, this allows for a primitive form of rendering (projection, shading and occlusion units).
These units do not even have learnable parameters.
Their purpose instead is to free the NN from learning basic concepts like occlusion, perspective, etc.

\refFigStart{NNInterpolation} shows interpolation of colors over space.
Consider regression of appearance using a multi-layer perceptron (MLP) \cite{oechsle2019texturefields,sitzmann2019srns,chen2019implicit} or convolutional neural network (CNN).
CNNs without the coord-conv trick \cite{liu2018coordconv} are particularly weak at such spatially-conditioned generation.
But even with coord-conv, this complex function is unnecessarily hard to find and slow to fit.

In contrast, methods that sample the observations using warping \cite{jaderberg2015spatial}
are much more effective to change the view \cite{zhou2016appearance}.
\revision{\refFigStart{Epipolar} shows a validation experiment, that compares classic pixel-basis interpolation and neural interpolation of color and warping.
Using a NN provides smooth epipolar lines, using warping, adds the details.
}
We will now detail our work, motivated by those observations.

\mycfigure{Overview}{
\revision{%
Data flow for an example with three dimensions (one view, one light, one temporal) and three samples, denoted as colors, as in \refFig{Teaser} and stacked vertically in each column.
In the first row, the 2$\times$3 Jacobian matrix is always visualized as separate channels \ie as three columns with two dimensions each.
Values are 2D-vectors, hence visualized as false colors.
At test time, the Jacobians are evaluated at the output \name\ coordinate only; hence, only a single row is shown.
In the second row, each observation is separately warped for shading and albedo, leading to 2$\times$3 flow, result and weight images.
The last row shows the flow of information as a diagram.
\emph{Learned} is a tunable,
\emph{Fixed} a non-tunable step (\ie without learnable parameters). 
\emph{Data} denotes access to inputs.
}
}

\mysection{Our Approach}{OurApproach}

We will first give a definition of the function we learn, followed by the architecture we choose for implementing it.

\mysubsection{Objective}{Objective}
We represent the \name as a non-linear function: \[
\xFieldOut^{(\parameters)}(\xCoord)
\in
\xCoords
\rightarrow
\mathbb R^{3\times\numberOfPixels},
\]
with trainable parameters $\parameters$
to map from an $\dimension$-dimensional \name coordinate $\xCoord\in\xCoords\subset\mathbb R^\dimension$ 
to 2D RGB images with $\numberOfPixels$ pixels.
The \name dimension depends on the capture modality:
A 4D example would be two spatial coordinates, one temporal dimension and one light angle.
Parametrization can also be as simple as scalar 1D time for video interpolation.
The symbol $\xFieldOut$ is chosen as images are in units of radiance with a subscript to denote them as output.

We denote as $\sparseXCoords\subset\xCoords$ the subset of \emph{observed} \name coordinates for which an image  $\xFieldIn(\sparseXCoord)$ was captured at the known coordinate $\sparseXCoord$.
Typically $|\sparseXCoords|$ is sparse, \ie small, like 3$\times$3, 5$\times$5 for view changes or even 2$\times$1 for stereo magnification.
We find this mapping $\xFieldOut$ by optimizing for
\[
\parameters=
\argmin{\parameters'}
\expected_{\sparseXCoord\sim\sparseXCoords}
||
\xFieldOut^{(\parameters')}(\sparseXCoord)
-
\xFieldIn(\sparseXCoord)
||_1,
\] where $\expected_{\sparseXCoord\sim\sparseXCoords}$ is the expected value across all the discrete and sparse \name coordinates $\sparseXCoords$.
In prose, we train an architecture $\xFieldOut$ to map vectors $\sparseXCoord$ to captured images  $\xFieldIn(\sparseXCoord)$ in the hope of also getting plausible images $\xFieldOut(\xCoord)$ for unobserved vectors $\xCoord$.
We aim for interpolation; $\xCoords$ is a convex combination of $\sparseXCoords$ and does not extend beyond.

Note that training never evaluates any \name coordinate $\xCoord$ that is not in $\sparseXCoords$, as we would not know what the image $\xFieldIn(\xCoord)$ at that coordinate would be.


\mysubsection{Architecture}{Architecture}
We model $\xFieldOut$ using three main ideas.
First, appearance is a combination of appearance in observed images.
Second, appearance is assumed to be a product of shading and albedo.
Third, we assume the unobserved shading and albedo at $\xCoord$ to be a warped version of the observed shading and albedo at $\sparseXCoord$.
These assumptions do not strictly need to hold, in particular not for splitting albedo and shading: when they are not fulfilled, the NN just has a harder time capturing the relationship of coordinates and images. 

Our pipeline $\xFieldOut$, depicted in \refFig{Overview}, implements this in four steps:
decoupling shading and albedo (\refSec{Delight}),
interpolating images (\refSec{Interpolation}) as a weighted combination of warped images (\refSec{Warping}),
representing flow using a NN (\refSec{Flow}) and
resolving inconsistencies (\refSec{Consistency}).

\mysubsubsection{De-light}{Delight}
De-lighting splits appearance into a combination of shading, which moves in one way in response to changes in \name coordinates, \eg highlights move in response to view changes or shadows move with respect to light changes, and albedo, which is attached to the surface and will move with geometry, \ie textures.

To this end, every observed image is decomposed as $
\xFieldIn(\sparseXCoord)=
\shading(\sparseXCoord)
\odot
\albedo(\sparseXCoord)
$, a per-pixel (Hadamard) product $\odot$ of a shading image $\shading$ and an albedo image $\albedo$.
This is done by adding one parameter to $\parameters$ for every observed pixel channel in $\shading$, and computing $\albedo$ from $\xFieldIn$ by division as $
\shading(\sparseXCoord)
=
\xFieldIn(\sparseXCoord)
\odot
\albedo(\sparseXCoord)^{-1}
$.
Both shading and albedo are interpolated independently:
\begin{equation}
\xFieldOut(\xCoord)
=
\interpolate(
    \albedo(\xFieldIn(\sparseXCoord)), \sparseXCoord\rightarrow\xCoord) 
\
\odot
\interpolate(
    \shading(\xFieldIn(\sparseXCoord)), \sparseXCoord\rightarrow\xCoord)
\end{equation}
and recombined into new radiance at an unobserved location $\xCoord$ by multiplication.
We will detail the operator $\interpolate$, working the same way on both shading $\shading(\xFieldIn)$ and albedo $\albedo(\xFieldIn)$, next.

\mysubsubsection{Interpolation}{Interpolation}
Interpolation warps all observed images and merges the individual results.
Both warp and merge are performed completely identically for shading $\shading$ and albedo $\albedo$, which we neutrally denote $I$, as in:
\begin{align}
\label{eq:Spatial}
\interpolate(I, \sparseXCoord\rightarrow\xCoord)
=
\sum_{\sparseXCoord\in
\sparseXCoords}
\left(
\consistency(\sparseXCoord
\rightarrow
\xCoord)
\odot
\warp(
I(\sparseXCoord), 
\sparseXCoord
\rightarrow
\xCoord)
\right)
.
\end{align}
The result is a weighted combination of deformed images.
Warping (\refSec{Warping}) models how an image changes when \name coordinates change by deforming it, and a per-pixel weight is given to this result to handle flow consistency (\refSec{Consistency}).

\mysubsubsection{Warping}{Warping}
Warping deforms an observed into an unobserved image, conditioned on the observed and the unobserved \name coordinates:
\begin{equation}
    \warp(\slice, \sparseXCoord\rightarrow\xCoord)
    \in
    \slices\times
    \xCoords\times
    \sparseXCoords
    \rightarrow
    \slices.
\end{equation}
We use a spatial transformer (STN) \cite{jaderberg2015spatial} with bi-linear filtering, \ie a component that computes all pixels in one image by reading them from  another image according to a given flow map.
STNs are differentiable, do not have any learnable parameters and are efficient to execute at test time.
The key question is, (\refFig{ImplicitWarp}) from which position $\warpedPixelPosition$ should a pixel at position $\pixelPosition$ read when the image at $\xCoord$ is reconstructed from the one at $\sparseXCoord$?

\myfigure{ImplicitWarp}{Implicit maps: implicit fields \textbf{(left)} typically use an MLP to map 3D position to color, occupancy etc.
We \textbf{(right)} add an indirection and map pixel position to texture coordinates to look up another image.}

To answer this question, we look at the Jacobians of the mapping from \name coordinates to pixel positions.
Here, Jacobians capture, for example, how a pixel moves in a certain view and light if time is changed, or its motion for one light, time and view coordinate if light is moved, and so forth.
Formally, for a specific pixel $\pixelPosition$, the Jacobian is:
\begin{equation}
\label{eq:Flow}
\flow(\xCoord)[\pixelPosition]
=
\frac
{\partial \pixelPosition (\xCoord)}
{\partial \xCoord}
\in
\xCoords
\rightarrow
\mathbb R^{2\times\dimension},
\end{equation}
where $[\cdot]$ denotes indexing into a discrete pixel array.
This is a Jacobian matrix with size $2\times\dimension$, which holds all partial derivatives of the two image pixel coordinate dimensions (horizontal and vertical) with  respect to all $\dimension$-dimensional \name coordinates.
A Jacobian is only differential and does not yet define the finite position $\warpedPixelPosition$ to read for at a pixel position $\pixelPosition$ as required by the STN.

To find $\warpedPixelPosition$ we will now \emph{project} the change in \name coordinate $\sparseXCoord\rightarrow\xCoord$ to 2D pixel motion using finite differences: \begin{equation}
\label{eq:PixelFlow}
\flowPix(\mathbf y\rightarrow\mathbf x)
[\pixelPosition]
=
\pixelPosition+
\Delta(\mathbf y\rightarrow\mathbf x) 
\flow(\xCoord)[\pixelPosition]
=
\warpedPixelPosition
.
\end{equation}
Here, the finite delta in \name coordinates $(\mathbf y\rightarrow\mathbf x)$, an $\dimension$-dimensio-nal vector, is multiplied with an $\dimension\times 2$ matrix, and added to the start position $\pixelPosition$, producing an absolute pixel position $\warpedPixelPosition$ used by the STN to perform the warp.
In other words, \refEq{Flow} specifies how pixels move for an infinitesimal change of \name coordinates, while \refEq{PixelFlow} gives a finite pixel motion for a finite change of \name coordinates.
We will now look into a learned representation of the Jacobian, $\flow$, the core of our approach. 

\mysubsubsection{Flow}{Flow}
Input to the flow computation is only the \name coordinate $\xCoord$ and output is the Jacobian (\refEq{Flow}).
We implement this function using a CNN, in particular.

\paragraph{Implementation}
Our implementation starts with a fully connected operation that transforms the coordinate $\xCoord$ into a 2$\times$2 image with 128 channels.
The coord-conv \cite{liu2018coordconv} information (the complete  $\xCoord$ at every pixel) is added at that stage.
This is followed by as many steps as it takes to arrive at the output resolution, reducing the number of channels to produce at $\dimension$ output channels.
For some input, it can be acceptable to produce a flow map at a resolution lower than the image resolution and warp high-resolution images using low-resolution flow, which preserves details in color, but not in motion.

\paragraph{Compression}
Changes in some \name dimension can only change the pixel coordinates in a limited way.
One example is view: all changes of pixel motion with respect to known camera motion can be explained by disparity \cite{forsyth2002computer}.
So instead of modeling a full 2D motion to depend on all view parameters, we only generate per-pixel disparity and compute the flow Jacobian from disparity in closed form using reprojection.
For our data, this is only applicable to depth, as no such constraints are in place for derivatives of time or light.

\paragraph{Discussion}
It should also be noted that no pixel-basis RGB observation
$\xFieldIn(\sparseXCoord)$ ever is input to $\flow$, and hence, all geometric structure is encoded in the network.
Recalling \refSec{Background}, we see this as both a burden, but also required to achieve the desired  interpolation property: if the geometry NN can explain the observations at a few $\sparseXCoord$, it can explain their interpolation at all $\xCoord$.
This also justifies why $\flow$ is a NN and we do not directly learn a pixel-basis depth-motion map: it would not be interpolatable.

\change{
An apparent alternative would be to learn $\flow'(\xCoord,\sparseXCoord)$ to depend on both $\sparseXCoord$ and $\xCoord$,
so as not to use a Jacobian, but allow any mapping.
Regrettably, this does not result in interpolation.
Consider a 1D view alone: Using $\flow(\xCoord)$ has to commit to one value that just minimizes image error after soft blending.
If a hypothetical $\flow'(\xCoord,\sparseXCoord)$ can pick any different value for every pair $\xCoord$ and $\sparseXCoord$, it will do so without \revision{incentive for a solution that is valid in between them}.
}

\change{Finally, it should be noted}, that using skip connections is not applicable to our setting, as the decoder input is a mere three numbers without any spatial meaning.

\mysubsubsection{Consistency}{Consistency}
To combine all observed images warped to the unobserved \name coordinate, we weight each image pixel by its \emph{flow consistency}.
For a pixel $\warpedPixelPosition$ to contribute to the image at $\pixelPosition$, the flow at $\warpedPixelPosition$ has to map back to $\pixelPosition$.
If not, evidence for not being an occlusion is missing and the pixel needs to be weighted down.

Formally, consistency of one pixel $\pixelPosition$ when warped to coordinate $\xCoord$ from $\sparseXCoord$ is the partition of unity of a weight function:
\begin{align}
\consistency(\sparseXCoord\rightarrow\xCoord)[\pixelPosition]=
\consistencyWeight(\sparseXCoord\rightarrow\xCoord)[\pixelPosition]
(
\sum_{\sparseXCoord'\in\sparseXCoords}
w(\sparseXCoord'\rightarrow\xCoord)[\pixelPosition]
)
^{-1}.
\end{align}
The weights $\consistencyWeight$ are smoothly decreasing functions of the $1$-norm of the delta of the pixel position $\pixelPosition$ and the backward flow at the position $\warpedPixelPosition$ where $\pixelPosition$ was warped to: 
\begin{align}
\consistencyWeight
(\sparseXCoord\rightarrow\xCoord)
[\pixelPosition]=
\exp(-\bandwidth|
\pixelPosition
-
\flowPix
(\xCoord\rightarrow\sparseXCoord)
[\warpedPixelPosition]
)|_1).
\end{align}
Here $\bandwidth=10$ is a bandwidth parameter chosen manually.
No benefit was observed when making $\bandwidth$ a vector or learning it.

\paragraph{Discussion}
In other work, consistency has been used in a loss, asking for consistent flow for unsupervised depth \cite{godard2017monodepth,zhou2017unsupervised} and motion \cite{zou2018df} estimation.
Our approach does not have consistency in the loss during training, but inserts it into the image compositing of the architecture, \ie also to be applied at test time.
In other approaches\textemdash that aim to produce depth, not images\textemdash consistency is not used at test time.
Our flow can be, and for our problem has to be inconsistent: for very sparse images such as three views, many occlusions occur, leading to inconsistencies.
Also flow due to, \eg caustics or shadows probably has a fundamentally different structure compared to multi-view flow, that has been not explored in the literature we are aware of.

The graphics question answered here is, however, what to do with inconsistencies.
To this end, instead of a consistency loss, we allow the architecture to apply multiple flows, such that the combined result is plausible when weighting down inconsistencies.
In the worst case, no flow is consistent with any other and $\consistencyWeight$ has similar but small values for large $\consistency$ which lead to equal weights after normalization, \ie linear blending.

\mysection{Results}{Results}
Here we will provide a comparison to other work (\refSec{ComparisonResults}), evaluation of scalability (\refSec{EvaluationResults}), and a discussion of applications (\refSec{ApplicationResults}).

Please see the supplemental materials for an interactive WebGL demo to explore different \name data sets using our method, \change{as well as a supplemental video to document temporal coherence}.

\begin{table*}[h]
\setlength{\tabcolsep}{1.4pt}
    \centering
    \caption{%
    Results of different methods \textbf{(columns)} for different dimensions \textbf{(rows)} according to different metrics.
    Below, the same data as diagrams.
    Colors encode methods.
    The best method according to one metric for one class of \name is denoted in bold font
    (for $L_2$ and VGG less is better, for SSIM more is better). \
    $^1$For view-time interpolation, combined with LLFF.
    }
    \begin{tabular}{ccc rrrrrrrrrr rrrrrrrrrr rrrrrrrrrr}
        \multicolumn{1}{c}{\multirow{2}{*}{\adjustbox{angle=90}{View}}}&
        \multicolumn{1}{c}{\multirow{2}{*}{\adjustbox{angle=90}{Time}}}&
        \multicolumn{1}{c}{\multirow{2}{*}{\adjustbox{angle=90}{Light}}}&
        \resultMethod{Linear}{44ABE0}&
        \resultMethod{Warping}{65BE67}&
         \resultMethod{Kalantari}{FFDE17}&
         \resultMethod{LLFF}{E07048}&
         \resultMethod{SuSloMo$^1$}{DB467F}&
         \resultMethod{NoWarp}{D1D3D4}&
         \resultMethod{NoCC}{808285}&
         \resultMethod{NoCons}{58595B}&
         \resultMethod{Ours}{000000}
        \\
        \cmidrule(lr){4-6}
        \cmidrule(lr){7-9}
        \cmidrule(lr){10-12}
        \cmidrule(lr){13-15}
        \cmidrule(lr){16-18}
        \cmidrule(lr){19-21}
        \cmidrule(lr){22-24}
        \cmidrule(lr){25-27}
        \cmidrule(lr){28-30}
        &&&%
        \multicolumn{1}{c}{\tiny VGG}&
        \multicolumn{1}{c}{\tiny MSE}&
        \multicolumn{1}{c}{\tiny SSIM}&
        \multicolumn{1}{c}{\tiny VGG}&
        \multicolumn{1}{c}{\tiny MSE}&
        \multicolumn{1}{c}{\tiny SSIM}&
        \multicolumn{1}{c}{\tiny VGG}&
        \multicolumn{1}{c}{\tiny MSE}&
        \multicolumn{1}{c}{\tiny SSIM}&
        \multicolumn{1}{c}{\tiny VGG}&
        \multicolumn{1}{c}{\tiny MSE}&
        \multicolumn{1}{c}{\tiny SSIM}&
        \multicolumn{1}{c}{\tiny VGG}&
        \multicolumn{1}{c}{\tiny MSE}&
        \multicolumn{1}{c}{\tiny SSIM}&
        \multicolumn{1}{c}{\tiny VGG}&
        \multicolumn{1}{c}{\tiny MSE}&
        \multicolumn{1}{c}{\tiny SSIM}&
        \multicolumn{1}{c}{\tiny VGG}&
        \multicolumn{1}{c}{\tiny MSE}&
        \multicolumn{1}{c}{\tiny SSIM}&
        \multicolumn{1}{c}{\tiny VGG}&
        \multicolumn{1}{c}{\tiny MSE}&
        \multicolumn{1}{c}{\tiny SSIM}&
        \multicolumn{1}{c}{\tiny VGG}&
        \multicolumn{1}{c}{\tiny MSE}&
        \multicolumn{1}{c}{\tiny SSIM}
        \\
        \toprule 
        \cmark&&&
        421 & 221 & .662 & 210 & 2.28 & 0.929 & 351 & 20.39 & .769 & 223 & 2.78 & .919 & \ndcell & \ndcell & \ndcell & 330 & 11.78 & .768 & 421 & 6.45 & .806 & 175 & 2.25 & .941 & \textbf{151} & \textbf{1.79} & \textbf{.951} \\
        &\cmark&&
        359 & 71 & .723 & \ndcell & \ndcell & \ndcell & \ndcell & \ndcell & \ndcell & \ndcell & \ndcell & \ndcell & 224 & 3.90 & .867 & 315 & 5.43 & .778 & 497 & 7.63 & .706 & 147 & 1.45 & .935 & \textbf{147} & \textbf{1.46} & \textbf{.935}
        \\
        &&\cmark&
        116 & 9 & .940 & \ndcell & \ndcell & \ndcell & \ndcell & \ndcell & \ndcell & \ndcell & \ndcell & \ndcell & 120 & .784 & .947 & 119 & 0.95 & .941 & 302 & 5.25 & .848 & 111 & 0.68 & .948 & \textbf{111} & \textbf{0.66} & \textbf{.948} 
        \\
        \midrule
        \cmark&\cmark&&
        620 & 176 & .558 & \ndcell & \ndcell & \ndcell & \ndcell & \ndcell & \ndcell & \ndcell & \ndcell & \ndcell & 269 & 1.99 & .892 & 388 & 7.67 & .775 & 571 & 14.97 & .632 & 273 & 2.61 & .888 & \textbf{252} & \textbf{2.00} & \textbf{.896}
        \\
        \midrule
        \cmark&\cmark&\cmark&
        522 & 209 & .584 & \ndcell & \ndcell & \ndcell & \ndcell & \ndcell & \ndcell & \ndcell & \ndcell & \ndcell & \ndcell & \ndcell & \ndcell & 523 & 20.60 & .595 & 493 & 10.10 & .692 & 419 & 7.09 &
        .719 & \textbf{247} & \textbf{2.19} & \textbf{.827}\\
        \bottomrule
    \end{tabular}
    \includegraphics[width=\linewidth]{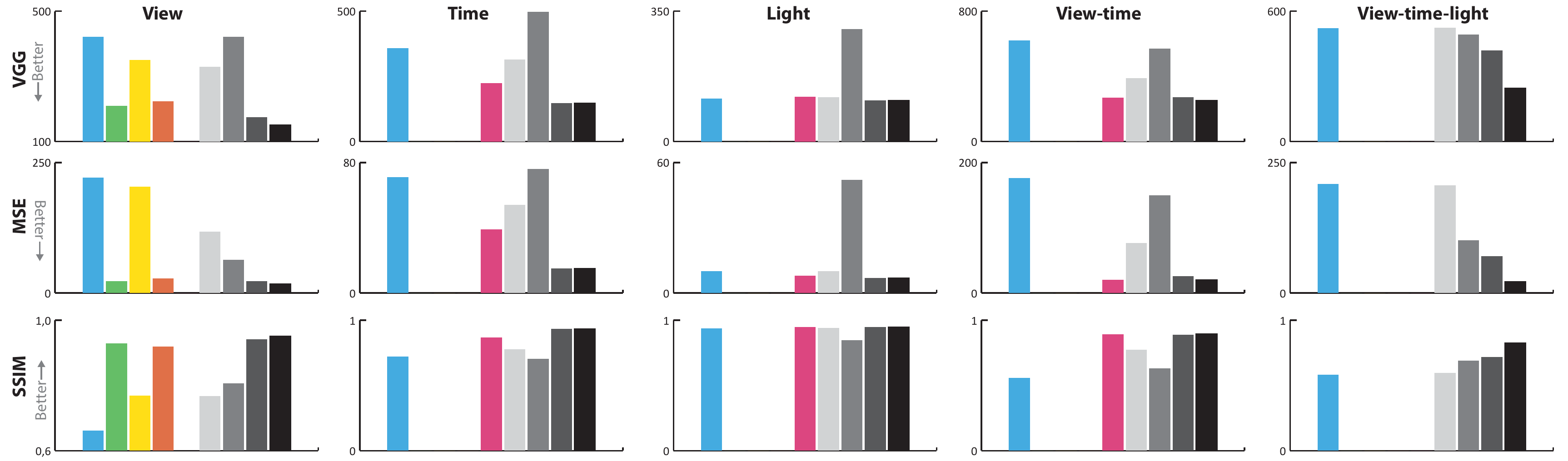}
\label{tbl:QuantitativeResults}
\end{table*}

\mycfigure{ViewComparison}{Comparison of our approach for view interpolation to other methods for two scenes \textbf{(rows)}.
The top scene, from \cite{Kalantari2016} is a dense LF; the one below, from \cite{penner2017soft}, is sparse.
Columns show, left to right, \textsc{Ours} at the position of the withheld reference, the results from (\textsc{Warping}, \textsc{KalantariEtAl}, \textsc{LLFF}, and \textsc{NoConsistency} and \textsc{Ours}), as well as the ground truth as insets.}

\mycfigure{TimeComparison}{Temporal interpolation for two scenes \textbf{(rows)} using different methods \textbf{(columns)}.
See \refSec{ComparisonResults} for a discussion.}

\mycfigure{SpaceTimeComparison}{Results for view-time interpolation.
The input was a 2$\times$2$\times2$ \name: 2$\times$2 sparse view observations with two frames.}

\mycfigure{ViewTimeLight}{Exploring a view-time-light \name.
In each column, we show a change in dominant \name dimension.
The input was a $3\times$3$\times3$  \name.
All images are at unobserved intermediate coordinates.
Colored arrows indicate how image features have moved in response.
}

\mysubsection{Comparison}{ComparisonResults}
We compare our approach to other \emph{methods}, following a specific \emph{protocol} and by different \emph{metrics} to be explained now:

\paragraph{Methods}
We consider the following methods:
\textsc{Ours},
\textsc{Blending},
\textsc{Warping},
\textsc{KalantariEtAl},
Local light-field fusion (\textsc{LLFF}),
\textsc{SuperSlowMo},
and three ablations of our approach:
\textsc{NoCordConv}, as well as
\textsc{NoWarping} and
\textsc{NoConsistency}.

Linear \textsc{Blending} is not a serious method, but documents the sparsity: plagued by ghosting for small baselines, we see our baseline/sparsity poses a difficult interpolation task, far from linear.
It is applicable to all dimensions.

\textsc{Warping} and \textsc{SuperSlowMo} first estimate the correspondence in image pairs \cite{sun2018pwc} or light field data \cite{dabala2016lightfield} and later apply warping \cite{mark1997post} with ULR-style weights \cite{buehler2001unstructured}.
Note how ULR weighting accounts for occlusion.
Warping is applicable to time (\namecite{jiang2018super}) and view interpolation (\namecite{dabala2016lightfield}).

\textsc{KalantariEtAl} and \textsc{LLFF} are the publicly available implementations of \namecite{Kalantari2016} and \namecite{Mildenhall2019}.
Both are applicable to and tested on lightfields, \ie view interpolation, only.

To evaluate other work in higher dimensions, we further explore their hypothetical combinations, such as first using \textsc{LLFF} for view interpolation followed by \textsc{SuperSlowMo} for time interpolation.

Finally, we compare three ablations of our method.
The first, \textsc{NoCordConv}, regresses without coord-conv, \ie will produce spatially invariant fields.
The second, \textsc{NoWarping}, uses direct regression of color values without warping.
The third, \textsc{NoConsistency}, does not perform occlusion reasoning but averages directly.
These are applicable to all dimensions.

As we did not have access to the reference implementation of Soft3D \cite{penner2017soft}, we test on their data and encourage qualitative comparison by inspecting our results in \refFig{ViewComparison} and their supplemental video.

\paragraph{Protocol}
Success is quantified as the expected ability of a method to predict a set of held-out LF observed coordinates $\heldOutXCoords$ when trained on $\sparseXCoords-\heldOutXCoords$, \ie $\expected_{\heldOutXCoord\sim\heldOutXCoords}\xFieldOut(\heldOutXCoord)\ominus_\mathrm m\xFieldIn(\heldOutXCoord)$, where $\ominus_\mathrm m$ is one of the metrics to be defined below.

For dense LF the held-out protocol follows \namecite{Kalantari2016}: four corner views as an input.
Sparse LF interpolation is on 5$\times$5, holding out the center one.
For time, interpolation triplets are used, \ie we train on past and future frames, withholding the middle one.

\paragraph{Metrics}
For comparing the predicted to the held-out view we use \change{the $L_2$, SSIM and VGG \revision{\cite{zhang2018unreasonable}} metrics}.

\paragraph{Data}
We use the publicly available LF data from \cite{levoy1996light}, \cite{penner2017soft}, \cite{dabala2016lightfield}, and \cite{Kalantari2016}, LF video data from \cite{Sabater2017}, sequences from \cite{Butler:ECCV:2012}, \change{relighting data from \namecite{xu2018deeprelighting}} as well as custom captured reflectance field video.
For aggregate statistics, we use 5 LFs, three videos and one view-time-light \name. 

Our own data is captured using a minimalist setup: a pair of mobile phones.
The first image takes the photo; the second one provides the light source.
Both are moved with one, two or three degrees of freedom, depending on the scene.
All animation is produced by stop motion.
We have captured several \names, but only one that has additional reference views to quantify quality.

\paragraph{Results}
\refTbl{QuantitativeResults} summarizes the outcome of the main comparison.
We see that our method provides the best quality in all tasks according to all metrics on all domains.

For example images corresponding to the plots in \refTbl{QuantitativeResults}, please see \refFig{ViewComparison} for interpolation in space, \refFig{TimeComparison} and \refFig{TimeWithFramesComparison} for time, \refFig{LightComparison} for light, \refFig{SpaceTimeComparison} for space-time and \refFig{ViewTimeLight} for view-time-light results.
In each figure we document the input view and multiple insets that show the results from all competing methods.

\refFigStart{ViewComparison} shows results for view interpolation.
Here, \textsc{Warping} produces crisp images, but pixel-level outliers that are distracting in motion, \eg for the bench.
\textsc{KalantariEtAl} and \textsc{LLFF} do not capture the tip of the grass (top row).
Instead, ghosted copies are observed.
\textsc{KalantariEtAl} is not supposed to work for larger baselines \cite{Kalantari2016} and only shown for completeness on the bench scene.
\textsc{LLFF} produces slightly blurrier results for the sparse bench scene.
Our \textsc{NoConsistency} shows the tip of the grass, but on top of ghosting.
\textsc{Ours} has details, plausible motion and is generally most similar to the ground truth.

The temporal interpolation comparison in \refFig{TimeComparison} indicates similar conclusions: \textsc{Blending} is not a usable option; not handling occlusion, also in time, creates ghosting due to overlap.
\textsc{SuperSlowMo} fails for both scenes as the motion is large.
The motion size can be seen from the linear blending.
Ultimately, \textsc{Ours} is similar to the ground truth.
The motion smoothness is best seen in the slow-mo application of the supplemental video.

Interpolation between triplets of images can represent strong, non-rigid changes involving transparency, scattering, etc. (\refFig{TimeWithFramesComparison}).

\myfigure{TimeWithFramesComparison}{Interpolation of two frames \textbf{(shown left)} compared to a reference using our approach and state-of-the-art SuperSlowMo \cite{jiang2018super}.}

\myfigure{LightComparison}{Interpolation in the light dimension. We note that the image is plausible, even in the presence of cast shadows or caustics and transparency, maybe at the slight expense of blurring highlights and ghosting shadows.}

Interpolation across light is seen in \refFig{LightComparison}.

\change{
For light interpolation, \namecite{xu2018deeprelighting} is an extension of our ablation \textsc{Direct}, by an additional optimization over sample placement when assuming a capture dome.
We will here compare to their implementation on their data.
Please note that their method cannot be applied to our data as it requires a custom capture setup.
\refFigStart{XuComparison} shows a comparison from interpolating across a neighborhood of 3$\times$3 images out of the 541 dome images, covering a baseline of approximately 20 degrees.
We see that direct regression blurs both the shadows and the highlights, while our method deforms the image, retaining sharpness.
\refTbl{XuComparison} quantifies this result as the average across their test images ``Dinosaur'', ``Jewel'' and ``Angel''. 
Besides the 10-degree column corresponding to \refFig{XuComparison}, we also include other baselines.
We see that for wider baselines, \namecite{xu2018deeprelighting} both methods converge in quality.
}

\myfigure{XuComparison}{Comparison between \namecite{xu2018deeprelighting} \textbf{(top)}, the GT \textbf{(middle)} and our approach \textbf{(bottom)} for a 10 degree baseline.}

\begin{table}[h]
    \setlength{\tabcolsep}{1.8pt}
    \centering
    \caption{Relighting comparison to \namecite{xu2018deeprelighting} for different baselines.}%
    \vspace{-0.25cm}%
    \begin{tabular}{l rrr rrr rrr}
        \multicolumn{1}{c}{\multirow{2}{*}{Method}}&
        \multicolumn{3}{c}{20$^{\circ}$ Baseline}&
        \multicolumn{3}{c}{30$^{\circ}$ Baseline}&
        \multicolumn{3}{c}{45$^{\circ}$ Baseline}
        \\
        \cmidrule(lr){2-4}
        \cmidrule(lr){5-7}
        \cmidrule(lr){8-10}
        &
        \multicolumn{1}{c}{\small VGG}&
        \multicolumn{1}{c}{\small MSE}&
        \multicolumn{1}{c}{\small SSIM}&
        \multicolumn{1}{c}{\small VGG}&
        \multicolumn{1}{c}{\small MSE}&
        \multicolumn{1}{c}{\small SSIM}&
        \multicolumn{1}{c}{\small VGG}&
        \multicolumn{1}{c}{\small MSE}&
        \multicolumn{1}{c}{\small SSIM}
        \\
        \toprule
        
        {\color[HTML]{D1D3D4}\ding{108}}
        \textsc{XuEtAl}
        &
        192&
        .1424&
        .954&
        194&
        .1561&
        .950&
        196&
        .1580&
        .950\\
        {\color[HTML]{000000}\ding{108}}
        \textsc{Ours}
        &
        93&
        .0335&
        .989&
        134&
        .0718&
        .970&
        169&
        .1220&
        .958\\
        \bottomrule
    \end{tabular}
    \label{tbl:XuComparison}
\end{table}

When interpolating across view and time as in \refFig{SpaceTimeComparison}, ghosting effects get stronger for others, as images get increasingly different.
\textsc{Ours} can have difficulties where deformations are not fully rigid, as seen for faces, but compensates for this to produce plausible images.

We conclude that both numerically and visually our approach can produce state-of-the-art interpolation in view and time in high spatial resolution and at high frame rates.
Next, we look into evaluating the dependency of this success on different factors.

\mysubsection{Evaluation}{EvaluationResults}
We evaluate our approach in terms of scalability with training effort and observation sparsity, speed and detail reproduction.
These tests are performed on the view interpolation only.

\myfigure{SeparationAnalysis}{Splitting albedo and shading: When the elephant's shadow meets a texture of \change{the Eiffel Tower} unprepared, a single-layer method such as \textsc{SuperSlowMo} cannot find a unique flow and produces artifacts.
Our approach leaves both shadow and texture structures mostly intact.}

\paragraph{Analysis of albedo splitting}
\refFigStart{SeparationAnalysis} shows an example of a scene that benefits from albedo splitting for a light interpolation.
We find that splitting albedo and shading is critical for shadows cast on textured surfaces.

\myfigure{Sparsity}{Visual quality of our approach as a function of increasing \textbf{(left to right)} training set size for view interpolation.}

\begin{wraptable}{r}{4cm}
    \caption{Error for the \emph{Crystal Ball} scene with resolution 512$\times$512 using different metrics \textbf{(columns)} for different view counts \textbf{(rows).}}%
    \label{tbl:SparsityEvaluation}
    \centering
    \begin{tabular}{cccc}         
         \multicolumn{1}{c}{LF}&
         \multicolumn{1}{c}{VGG19}&
         \multicolumn{1}{c}{L2}&
         \multicolumn{1}{c}{SSIM}\\
         \toprule
          3$\times$3&  140& .005& .90\\
          5$\times$5&  119& .003& .93\\
          9$\times$9&  102& .002& .95\\
          \bottomrule
    \end{tabular}
\end{wraptable}
\paragraph{Observation sparsity}
We interpolate from extremely sparse data.
In \refTbl{SparsityEvaluation} we evaluate the quality of our interpolation depending on the number of training exemplars, also seen in \refFig{Sparsity}.

\mycfigure{Iteration}{Progression of visual fidelity for different training efforts (horizontal axis) for two insets (vertical axis) in one scene.
After 500 epochs (ca.~30 minutes) the result is usable, and it converges after 1000 epochs (ca.~1h).
Note that epochs are short as we only have 5$\times$5 training examples.}

\paragraph{Speed}
At deployment, our method requires no more than taking a couple of numbers and passing them through a decoder for each observation, followed by warping and a weighting.
The end-speed for view navigation is around 20~Hz (on average 46\,ms per frame) at 1024$\times$1024 for a 5$\times$5 LF on an Nvidia 1080Ti with 12 GB RAM.

\begin{wraptable}{r}{4cm}
\setlength{\tabcolsep}{3pt}
    \centering
    \caption{Training time (minutes) and network parameters for different resolutions for a 5$\times$5 LF array and spatial interpolation.}
    \begin{tabular}{llcccc}     
     &
     512$^2$
     &
     1024$^2$
     &
     1764$^2$
     \\
     \toprule     
     Time
     & 
     28 & 60 & 172
     \\     
     Params& 
     482\,k& 492\,k& 492\,k\\    
     \bottomrule
\end{tabular}
\label{tbl:TrainingTime}
\end{wraptable}
\paragraph{Training effort}
Our approach needs to be trained again for every LF.
Typical training time is listed in \refTbl{TrainingTime}.
\refFigStart{Iteration} shows progression of interpolation quality over learning time.
We see that even after little training, results can be acceptable.
Overall, we see that training the NN requires a workable amount of time, compared approaches trained in the order of many hours or days.

\mywrapfigure{Smoothness}{0.45}{\change{Correspondence for \refFig{SpaceTimeComparison}.}}%

\paragraph{Smoothness}
The depth and flow map we produce are smooth in view and time and may lack detail. It would be easy to add skip connections to get the details from the appearance.
Regrettably, this would only work on the input image, and that needs to be withheld at training, and is unknown at test time.
An example of this is seen in \refFig{Smoothness}. This smoothness is one source of artifacts.
Overcoming this, \eg using an adversarial design, is left to future work.

\paragraph{\change{Coherence}}
\change{Visual coherence across dimensions, when traversing the X-space smoothly, is best assessed from the supplemental video.
Our method might miss details or over-smooth, but is coherent, as first, we never regress colors that flicker, only texture coordinates; second, Jacobians are multiplied with view differentials in a linear operation, and hence smooth; third as the NNs to produce Jacobians are smooth functions and, finally, soft occlusion is smooth.
}

\mysubsection{Applications}{ApplicationResults}
\refFigStart{Application} demonstrates motion blur (time interpolation), depth-of-field (view interpolation), and both (interpolating both).

\myfigure{Application}{Two LF video-enabled effects, computed using view interpolation: Depth-of-field \textbf{(left)} and motion blur \textbf{(right)}.
\change{For both, we generate and average many images at \name{} coordinates covering a lens resp. shutter.}}


\mysection{Discussion / Limitations}{Discussion}

We find the success of our method to largely depend on three factors, Data, Model and Capacity, which we will discuss next.
\change{Please also see the supplemental video and \refFig{FailureAnimation} for examples of such limitations.}

\myfigure{Failure}{
\change{Two failure cases of our method, documenting, left, insufficient data (the lamp post is only visible in one view and happens to become attached to the foreground leaf) and right, insufficiency of the capacity (the depth structure of the twigs is too complex to be represented by an architecture we can train at this points).
Please see the supplemental video for changes of view point to best appreciate the effect.}}

\mycfigure{FailureAnimation}{%
\revision{
Our interpolation results for two scenes (``Apple'' and ``Chair'') from the supplemental video:
Insets in red identify regions where artifacts appeared, and insets in green indicate challenging examples which our method  interpolated successfully.
Artifacts mainly happened due to lack of training data; in the apple scene \textbf{(top)}, a 3$\times$3$\times$3 \name\ capture, the caustics in the shadow shows in only one view, and the foam is stochastic and different at each level of the liquid.
In these regions, appearance does not properly interpolate but fades in and out, leading to ghosting or blurring, best seen in the supplemental video.
In the chair scene \textbf{(bottom)}, which is a  5$\times$5$\times$5 \name, the texture on the carpet beneath the chair gets blurry as this part of the carpet becomes visible only in one view due to occlusion caused by the chair and its shadow.
However, our method could handle soft shadow casting on a textured background or when there is a moving shadow of complex object occluded with the object itself. 
}}
\paragraph{Data}
We train from very sparse observations, often only a dozen images.
It is clear that information not present in any image, will not be reconstructed.
Even parts observed in only one image can be problematic \change{(\refFig{Failure}, left)}.
A classic example is occlusion: if only three different views are available and two occlude an area that is not occluded in one view, this area will be filled in.
However, this fill-in will occur in the domain we learn, the \name Jacobian.
Hence, disoccluded pixels will change their position similarly to their spatial neighbours.
Artifacts manifest as rubber-like stretches between the disoccluding and the occluding object.
\revision{The chair example from the supplemental video and \refFig{FailureAnimation} shows artifacts resulting from lack of data.
Similarly, the foam in the supplemental video and \refFig{FailureAnimation} is stochastic and different in every image, and hence unable to form fine-scale correspondence.}
The consistency weighting typically removes them.
Future work might overcome this limitation by training on more than one scene.

\paragraph{Model}
We combine a primitive, hard-wired image formation model with a learned scene representation.
As long as the data roughly follows this model, this is a winning combination. 
Scenes that are entirely beyond the model's scope might fail and will do so independently of the amount of data or the representation capacity.

Our key assumption is that changes are explained by flow.
This is not a reasonable assumption with dominant transparency \cite{Kopf2013}.
\revision{Changes in brightness due to casual capture with auto-exposure can cause variation that our deformation  model fails to explain.}
In an \name non-unique flow is common: after one bounce, multiple indirect shadows might overlap and move differently.

We address this by processing the  signal, so a unique flow becomes more applicable: by splitting shading and albedo, by representing the full \name Jacobian, by learning a non-linear inverse flow instead of linearly interpolating a forward flow, etc.
Finally, if all flows were wrong, consistency weighting degenerates to linear blending.
Future work could learn layered flow \cite{Sun2012}.

\paragraph{Capacity}
Finally, even if all data is available, the model is perfect and the model assumptions are fulfilled, the NN needs to have the capacity to represent the input to the model.
Naturally, any finite model can only be an approximation, and hence, the flow, and consequently shape, illumination and motion is smooth.
The NN allows for some level of sharpness via non-linearities as in other implicit representations \cite{niemeyer2019occupancy,oechsle2019texturefields,chen2019implicit} but the amount of information is finite \change{(\refFig{Failure}, right)}.
Capturing sharp silhouettes is clearly possible, but to represent a scene with stochastic variation, stochasticity should be inserted \cite{karras2019style} in combination with a style loss.

\mysection{Conclusion}{Conclusion}
We have demonstrated representing an \name as a NN that produces images, conditioned on view, time and light coordinates.
The interpolation is high-quality and high-performance, outperforming several competitors \change{for dynamic changes of advanced light transport (all BRDFs, (soft) shadows, GI, caustics, reflections, transparency), as well as fine spatial details (plant structures), both for single objects (still-life scenes) and entire scenes (tabletop soccer, parks)}.

The particular structure of a network that combines a learnable view-time geometry model, combined with warping and reasoning on consistency, has shown to perform better than direct regression of color or warping without handling occlusion and state-of-the-art domain-adapted solutions.

We want to reiterate that, partly, this success is possible because we changed a general task to a much simpler one: instead of interpolating all possible combinations of images, we only interpolate a fixed set.
Strong generalization is a worthwhile and exciting scientific goal, in particular from an AI perspective.
But, depending on the use case, it might not be required in applied graphics:
With our approach, after 20 minutes of pre-calculation, we can deploy an \name in a VR application to play back at interactive rates.
A user enjoying this high-quality visual experience might not ask if the same network could generalize to a different scene or not.

In future work, \change{other data such as data from Lightstages or sparse and unstructured capture, as well as extrapolation, should be explored.}
We aim to further reduce training time (eventually using learned gradient descent \cite{flynn2019deepview}), and explore interpolation along other domains such as wavelength or spatial audio \cite{engel2017neural}, as well as reconstruction from even sparser observations. 

\bibliographystyle{ACM-Reference-Format}

\end{document}